\documentclass[10pt,twocolumn,letterpaper]{article}

\usepackage{cvpr}
\usepackage{times}
\usepackage{epsfig}
\usepackage{graphicx}
\usepackage{amsmath}
\usepackage{amssymb}

\usepackage[breaklinks=true,bookmarks=false]{hyperref}

\cvprfinalcopy 

\def\cvprPaperID{****} 

\newcommand{\norm}[1]{\left\lVert#1\right\rVert}

\begin{document}

\title{Deep Multimodal Feature Analysis for Action Recognition in RGB+D Videos\\{\normalsize(Technical Report)}}

\author{Amir~Shahroudy$^{\dagger,\ddagger}$\\{\tt\small amir3@ntu.edu.sg}
	\and Tian-Tsong~Ng$^{\ddagger}$\\{\tt\small ttng@i2r.a-star.edu.sg}
	\and Yihong~Gong$^{\S}$\\{\tt\small ygong@mail.xjtu.edu.cn}
	\and Gang~Wang$^{{\dagger,1}}$\\{\tt\small wanggang@ntu.edu.sg}
	\and $\dagger$ School of Electrical and Electronic Engineering, Nanyang Technological University, Singapore
	\\$\S$ Institute of Artificial Intelligence and Robotics, Xi\'an Jiaotong University, Shaanxi, China
	\\$\ddagger$ Institute for Infocomm Research, Singapore
}

\maketitle

\begin{abstract}
Single modality action recognition on RGB or depth sequences has been extensively explored recently. 
It is generally accepted that each of these two modalities has different strengths and limitations for the task of action recognition. 
Therefore, analysis of the RGB+D videos can help us to better study the complementary properties of these two types of modalities and achieve higher levels of performance.   
In this paper, we propose a new deep autoencoder based shared-specific feature factorization network to separate input multimodal signals into a hierarchy of components. 
Further, based on the structure of the features, a structured sparsity learning machine is proposed which utilizes mixed norms to apply regularization within components and group selection between them for better classification performance.
Our experimental results show the effectiveness of our cross-modality feature analysis framework by achieving state-of-the-art accuracy for action classification on five challenging benchmark datasets.
\end{abstract}

\section{Introduction}
\footnotetext[1]{Gang Wang is the corresponding author.}
Recent development of range sensors had an indisputable impact on research and applications of machine vision.
Range sensors provide depth information of the scene and objects, which helps in solving problems that are considered hard for RGB inputs \cite{kinectSurvey2013}.

Human activity recognition is one of the active fields in computer vision and has been explored extensively. 
Recent advances in hand-crafted \cite{idt_ICCV2013,peng2014bag} and convnet-based \cite{twostreamCNN} feature extraction and analysis of RGB action videos achieved impressive performance. 
They generally recognize action classes based on appearance and motion patterns in videos. 
The major limitation of RGB sequences is the absence of 3D structure from the scene. 
Although some works are done towards this direction \cite{eigen2014depth}, recovering depth from RGB in general is an underdetermined problem. 
As a result, depth sequences provide an exclusive modality of information about the 3D structure of the scene, which suits the problem of activity analysis \cite{HON4D,HOPC,VemulapalliCVPR14,actionletPAMI,MMTW,xiaCVPR13spatio,Orderlet}. 
This complements the textural and appearance information from RGB. 
Our goal in this work is to analyze the multimodal RGB+D signals for identifying the strengths of respective modalities through teasing out their shared and modality-specific components and to utilize them for improving the classification of human actions.

Having multiple sources of information, one can find a new space of common components which can be more robust than any of the input features. 
Through linear projections, canonical correlation analysis (CCA) \cite{CCASurvey,hotelling1936relations} gives us the correlated form of input modalities which in essence is a robust representation of multimodal signals. 
However, the downside of CCA is the linearity limitation. 
Kernel canonical correlation analysis (KCCA) \cite{KCCA2000} extended this idea into nonlinear kernel-based projections, which is still limited to the representation capacity of the kernel's space and is not able to disentangle the high-level nonlinear complexities between the input modalities.
Further, the traditional solutions of CCA and KCCA are to solve the maximization of correlation between the projected vectors analytically, which does not scale well with the size of the data. 


To overcome these limitations, a new deep autoencoder-based nonlinear common component analysis network is proposed to discover the shared and informative components of input RGB+D signals.

Besides the shared components, each input modality has specific features which carry discriminative information for the recognition task.
In this respect, we can enhance the representation by incorporating the modality-specific components of respective modalities \cite{MVSV_CVPR14,salzmann2010factorized}.
Based on this intuition, at each layer our deep network factorizes the multimodal input features into their shared and modality-specific components.
By stacking such layers, we further decode the complex and highly nonlinear representations of the input modalities in a nonlinear fashion.

Across the layers, our deep multimodality analysis extracts a set of structured features which consist of hierarchically factorized multimodal components.
The common components are robust against noise and missing information between the modalities, and the modality-specific components carry the remaining informative features which are irrelevant to the other modality.
To effectively perform recognition tasks on our structured features, we design a structured sparsity-based learning framework.
With different mixed norms, features of each component can be grouped together and group selection can be applied to learn a better classifier.
We also show that the advantage of our learning framework is more significant as network gets deeper.

The contributions of this work are two-fold: first we introduce a new deep learning network for hierarchical shared-specific factorization of RGB+D features. 
Second, a structured sparsity learning machine is proposed to explore the structure of hierarchical factorized representations for effective action classification.

The rest of this paper is organized as follows. 
Section \ref{sec:relatedwork} explores the related work. 
Section \ref{sec:approach} introduces the proposed deep component factorization network. 
Section \ref{sec:classifier} describes our classification framework for factorized components. 
Section \ref{sec:exp} provides our experimental results, 
and section \ref{sec:conclusion} concludes the paper.

\section{Related work}
\label{sec:relatedwork}

There are other works which applied deep networks to multimodal learning.
The work in \cite{ngiam2011multimodal,JMLR:v15:srivastava14b} used DBM for finding a common space representation for two input modalities, and predict one modality from the other.
Andrew \etal \cite{DCCA} proposed a deep canonical correlation analysis network with two stacks of deep embedding followed by a CCA on top layer.
Our method is different from these works in two major aspects.
First, the previous work performed the multimodal analysis in just one layer of the deep network, but our proposed method performs the common component analysis in every single layer.
Second, we incorporate modality-specific components in each layer to maintain all the informative features, at each layer.

Jia \etal \cite{jia2010factorized} factorized the input features to shared and private components by applying structured sparsity, for the task of multi-view learning on human pose estimation, with linearity assumption. 
Cai \etal \cite{MVSV_CVPR14} proposed a nonlinear factorization of the features into common and individual components, towards a better representation of features for action recognition.
They utilized mixture models to add nonlinearity to linear probabilistic CCA \cite{bach2005probabilistic}. 
Our proposed technique stacks layers of nonlinear shared component analysis to progressively disentangle highly nonlinear correlations between the input features.

While learning frameworks in \cite{Wang_2013_ICCV,icml2013_wang13c,wang2013robust,6619242} applied structured sparsity for other similar tasks, our structured sparsity learning machine extends the sparse selection into two levels of concurrent component and layer selection, which is more suited to the hierarchical outputs from our deep factorization network.

Recent single modality action recognition methods on depth signals can be divided into two major groups: depth map analysis methods \cite{RangeSample,HON4D,HOPC_PAMI,xiaCVPR13spatio} and skeleton based methods \cite{Luo_2013_ICCV,VemulapalliCVPR14,skeletalQuads,actionletPAMI,MMTW}.

The first group extract the action descriptors directly from depth map sequences.
The idea of spatio-temporal interest points \cite{STIP} was applied in depth videos by \cite{xiaCVPR13spatio}.
They also proposed depth cuboid similarity features to represent local patches.
HON4D \cite{HON4D} represents depth sequences as histograms of 4D oriented normals of local patches, quantized on the vertices of a regular polychoron.
Rahmani \etal \cite{HOPC, HOPC_PAMI} achieved higher levels of robustness against viewpoint variations by using histograms of oriented principle components.
Lu \etal \cite{RangeSample} proposed binary range-sample descriptors based on $\tau$ tests on depth patches.
The work of \cite{cnn_for_depth_action_THMS} applied convolutional networks for learning action classes on depth maps.
Rahmani and Mian \cite{Rahmani_2015_CVPR} introduced a nonlinear knowledge transfer model to transform different views of human actions to a canonical view.

The second group of methods represent actions based on the 3D positions of major body joints, which are available for most of depth based action datasets.
Luo \etal \cite{Luo_2013_ICCV} proposed a novel skeleton-based discriminative dictionary learning method, utilizing group sparsity and geometry constraints.
Vemulapalli \etal \cite{VemulapalliCVPR14} represented skeletons as points and actions as curves in a Lie group using the 3D relative geometry between body parts. 
Evangelidis \etal \cite{skeletalQuads} proposed a compact and view-invariant representation of body poses calculated from joint positions.
Wang \etal \cite{7008115} introduced a mining technique to find part-based mid-level patterns (frequent local parts) and aggregated the local representations as bag-of-FLPs to be classified by a SVM.
Veeriah \etal \cite{diffRNN} extended the structure of the long short-term memory (LSTM) units \cite{lstm} to learn differential patterns in skeleton locations.
The work of \cite{rnnskeleton_cvpr15} introduced a hierarchy of recurrent networks to learn part-based motion patterns and combine them for action classification.
Zhu \etal \cite{cooccurrance} proposed a new regularization term for learning co-occurrences of motion patterns among different joint groups.
The work of \cite{Amir-Dataset-CVPR} introduced a new part-aware LSTM structure to discover the long-term motion patterns of skeleton-based body parts separately and learn the action classes based on these representations.
In \cite{actionletPAMI}, motion and local depth based appearance of each body joint was encoded using Fourier transform over a temporal pyramid.
They also proposed a mining method to find the set of most representative body joints for each action class.
Shahroudy \etal \cite{MMMP_PAMI} formulated the discriminative joint selection by introducing a hierarchical mixed norm.
The work of \cite{6836044} combined different spatio-temporal depth and skeleton based gradient features and applied a random decision forest for action classification.
Meng \etal \cite{7284883} proposed a real-time action recognition method by applying random forest classifier on a set of distance values between the body joints and interacted objects.
Wang and Wu \cite{MMTW} applied max-margin time warping to match the descriptors of skeletons over the temporal axis and learn phantom templates for each action class.
An extensive review of different approaches and techniques on 3D skeletal data is done by \cite{2016arXiv160101006H}.
The fusion of various depth based features is also studied by \cite{Yu_DepthFusion_ACM}.

Multimodality analysis of RGB+D action videos is studied by \cite{rgbdhudaact,RGGP,6918467,jianfang_CVPR15,Kong_2015_CVPR,6411953,depthinduced,Liu201579,6696669,Kosmopoulos2013,AmirAthens,RGGP}.
Ni \etal \cite{rgbdhudaact} introduced a RGB+D fusion method by concatenating depth descriptors to RGB based representations of STIP points.
Liu and Shao \cite{RGGP} introduced a genetic programming framework to improve the RGB and depth descriptors and their fusion simultaneously through an iterative evolution.
The work of \cite{6918467} solved the problem of RGB+D action recognition by utilizing RGB information for better tracking of interest point trajectories and describe them by depth-base local surface patches.
Hu \etal \cite{jianfang_CVPR15} proposed a heterogeneous multitask feature learning framework to mine shared and modality-specific RGB+D features.
The work of \cite{Kong_2015_CVPR} applied projection matrices to the common and independent spaces between RGB and depth modalities.
They learned their model by minimizing the rank of their proposed low-rank bilinear classifier.

The work of \cite{6411953} also extracted STIPs from RGB and combined their HOG and HOF descriptors from RGB channel with local depth patterns (LDP) features from depth channel to fuse the two modalities.
Depth-induced multiple channel STIPs \cite{depthinduced}, also added depth distances into GMM-based STIP representations.
In \cite{Liu201579} the STIP detection is done separately on RGB and depth and the HOGHOF descriptors are fused by combining the BOW representations of LLC codes of local features.
Tsai \etal \cite{6696669} used depth channel to segment the human body into known parts.
STIPs with descriptors on RGB and depth channels are aggregated for each part by BOF representation over temporal pyramids.
They assigned higher weights to non-occluded body parts to achieve a more robust global representations for action recognition.
Multistream fused hidden Markov model was utilized to fuse pixel change history feature from RGB with MHI feature from depth channel by \cite{Kosmopoulos2013}.
The work of \cite{AmirAthens} proposed a structured sparsity based fusion for RGB+D local descriptors.
An evolutionary programming RGB+D fusion method was proposed by \cite{RGGP}.
The proposed RGB+D analysis frameworks, are different from these methods, since our focus is on studying the correlation between the two modalities in the local level features and factorizing them to their correlated and independent components. 

Recent advances of visual recognition in digital images using deep convolutional networks \cite{alexnet, vggnet, googlenet, Gu2015arXivSurvey} also inspired the research in video analysis.
Ng \etal \cite{Ng_2015_CVPR} studied two techniques of feeding videos to convnets for video classification.
They proposed temporal pooling of the convnet-based features of frames to aggregate video descriptors from frame features.
They also studied the advantages of utilizing a long short-term memory \cite{lstm} network stacked over a convnet for video classification.
Simonyan and Zisserman \cite{twostreamCNN} fed a fixed length video sequence and its optical flow to a two-streamed convnet and fused the scores of the two streams in the end to classify the action labels.
Wang \etal \cite{Wang_2015_CVPR} combined the advantages of hand-crafted trajectory-based features and deep convnet learning based methods by applying \cite{twostreamCNN}'s network along the motion trajectories of input videos.
A novel deep convoultional framework for video event detection and evidence recounting was proposed by \cite{Gan_2015_CVPR}.
They introduced a back pass technique to localize the key evidences of the interested events in spatial-temporal domain.
Tran \etal \cite{Tran_2015_ICCV} studied the fully three dimensional convnet based video analysis and evaluated their proposed framework on various video analysis tasks including action recognition.

The applications of recurrent neural networks for 3D human action recognition were explored very recently.
Du \etal \cite{rnnskeleton_cvpr15} applied a hierarchical RNN to discover common 3D action patterns in a data-driven learning method.
They divided the input 3D human skeletal data to five groups of joints and fed them into a separated bidirectional RNN.
The output hidden representation of the first layer RNNs were concatenated to form upper-body and lower-body mid-level representation and these were fed to the next layer of bidirectional RNNs.
The holistic representation for the entire body was obtained by concatenating the output hidden representations of these second layer RNNs and it was fed to the last RNN layer.
The output hidden representation of the final RNN was fed to the softmax classifier for action classification.
Differential RNN \cite{diffRNN} added a new gating mechanism to the traditional LSTM to extract the derivatives of internal state (DoS).
The derived DoS was fed to the LSTM gates to learn salient dynamic patterns in 3D skeleton data.
The work of \cite{cooccurrance} introduced an internal dropout mechanism applied to LSTM gates for stronger regularization in the RNN-based 3D action learning network.
To further regularize the learning, a co-occurrence inducing norm was added to the network's cost function which enforced the learning to discover the groups of co-occurring and discriminative joints for better action recognition.
Liu \etal \cite{Liu_2016_ECCV, Liu_SUB_PAMI} extended the recurrent network based sequence analysis towards sequences of body joints.
They added a new dimension dimension to the structure of a deep LSTM-based framework to learn the features over time and over the sequences of joints concurrently.
To apply ConvNet-based learning to this domain, \cite{Rahmani_2016_CVPR} used synthetically generated data and fitted them to real mocap data.
Their learning method was able to recognize actions from novel poses and viewpoints.

Different from other methods, the proposed framework analyzes the components between the two modalities in a deep network, and factorizes the input RGB+D features into their shared and specific components in a hierarchy of nonlinear layers. 
Our solution is general and can be applied on any type of multimodal features to analyze their cross-modality components.

\section{Deep shared-specific component analysis}
\label{sec:approach}
\begin{figure}[t]
	\centering
	\def\svgwidth{\columnwidth}
	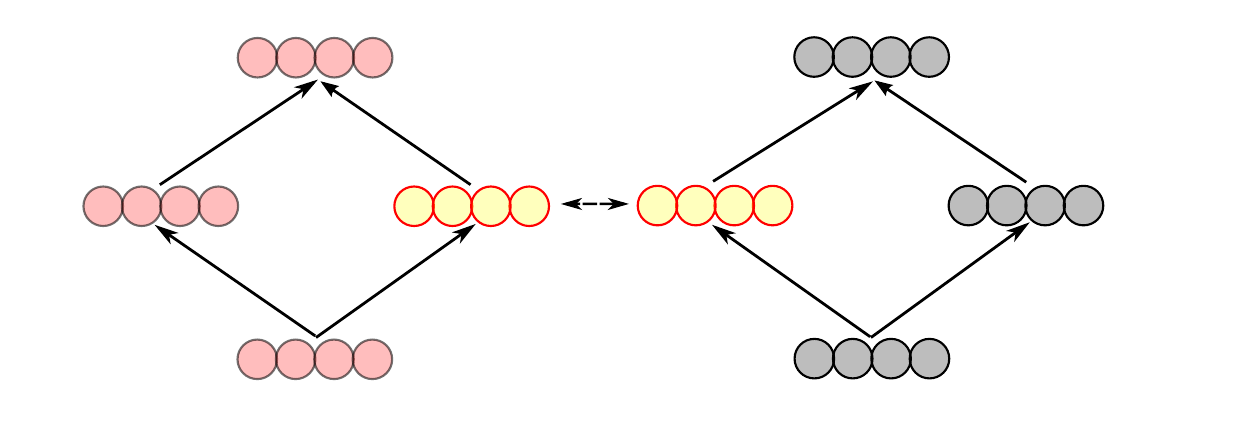
	\caption{Illustration of the proposed single layer shared-specific component analysis.
		${\bf X}_r$ and ${\bf X}_d$ are input RGB and depth based features. 
		We factorize each input feature into shared $({\bf Y})$ and specific $({\bf Z})$ components by forcing the {\bf Y} vectors to be close, and the input features to be reconstructible from derived components.
	}
	\label{fig:unit}
\end{figure}
We have two sets of features extracted from different modalities of data (RGB and depth signals) as our input for the task of action classification. 
State-of-the-art RGB based features \cite{dense_trajectories_IJCV,idt_ICCV2013} include 2D motion patterns and appearance information of objects and scenes. 
On the other hand, various depth-based features \cite{HON4D,HOPC,actionletPAMI,xiaCVPR13spatio} encode 3D shape and motion information, without appearance and texture details.
Consequently, it is beneficial to fuse the complementary RGB and depth-based features for better performance in action analysis.

There are different techniques for feature fusion. 
The choice of fusion strategy should rely on dependency of features.
When features have very high dependency, descriptor-level fusion gives the best outcome, and when multiple groups of features have very low interdependency, kernel-level fusion performs better 
\cite{salzmann2010factorized}. 
Since RGB and depth based features encode an entangled combination of common and modality-specific information of the observed action, they are neither independent nor fully correlated. 
Therefore, it is reasonable to embed the input data into a space of factorized common and modality-specific components.
The combination of the shared and specific components in input features can be very complex and highly nonlinear. 
To disentangle them, we stack layers of nonlinear autoencoder-based component factorization to form a deep shared-specific analysis network.

In this section, we first introduce our basic framework of shared-specific analysis for multimodal signal factorization, then describe the deep network of stacked layers, where each layer performs factorization analysis and collectively produce a hierarchical set of shared and modality-specific components.

\subsection{Single layer shared-specific analysis}
\label{sec:unit}
Let us notate input RGB features by ${\bf X}_r$ and depth features by ${\bf X}_d$. 
We propose to factorize each input feature pattern into two spaces: first, common component space which corresponds to the highest correlation with the other modality $({\bf Y}_r,{\bf Y}_d)$, and second, its modality-specific feature component space $({\bf Z}_r,{\bf Z}_d)$:
\begin{equation}
	\begin{bmatrix}
		{\bf Y}_r\\{\bf Y}_d\\{\bf Z}_r\\{\bf Z}_d
	\end{bmatrix}
	= g({\bf X}_r,{\bf X}_d;\Omega)
\end{equation}
where $\Omega$ is the set of model parameters that will be learned from the training data. 
We propose a sparse autoencoder-based network as the $g(.)$ function, as illustrated in \figurename{~\ref{fig:unit}}. 

Feature vectors of each modality are factorized into $\bf Y$ and $\bf Z$ which represent shared and individual components of each modality respectively. 
Each component is derived from a linear projection of the input features followed by a nonlinear activation. Mathematically:
\begin{eqnarray}
	\label{eqn:Yr}{\bf Y}_r&=&f({\bf W}_r {\bf X}_r+{\bf b}_{Y_r} {\bf 1}^n) \\
	\label{eqn:Zr}{\bf Z}_r&=&f({\bf V}_r {\bf X}_r+{\bf b}_{Z_r} {\bf 1}^n)
\end{eqnarray}
in which $f(.)$ is a nonlinear activation function. 
We use sigmoid scaling in our implementation. 
${\bf b}_{Y_r}$ and ${\bf b}_{Z_r}$ are bias terms. 
Similarly, for the depth based input, we have:
\begin{eqnarray}
	\label{eqn:Yd}{\bf Y}_d&=&f({\bf W}_d {\bf X}_d+{\bf b}_{Y_d} {\bf 1}^n) \\
	\label{eqn:Zd}{\bf Z}_d&=&f({\bf V}_d {\bf X}_d+{\bf b}_{Z_d} {\bf 1}^n)
\end{eqnarray}

To prevent output degeneration, we expect the original features to be reconstructible from their factorized components \cite{RICA_NIPS11}:
\begin{eqnarray}
	\label{eqn:Xtr}
	\widetilde{\bf X}_r&=&
	f(\begin{bmatrix}{\bf Q}_r~{\bf U}_r\end{bmatrix} \begin{bmatrix} {\bf Y}_r \\ {\bf Z}_r \end{bmatrix}+{\bf b}_{\widetilde{X}_r} {\bf 1}^n)\nonumber\\
	&=&f({\bf Q}_r{\bf Y}_r + {\bf U}_r{\bf Z}_r+{\bf b}_{\widetilde{X}_r} {\bf 1}^n) \\
	\label{eqn:Xtd}
	\widetilde{\bf X}_d&=&
	f(\begin{bmatrix}{\bf Q}_d~{\bf U}_d\end{bmatrix} \begin{bmatrix} {\bf Y}_d \\ {\bf Z}_d \end{bmatrix}+{\bf b}_{\widetilde{X}_d} {\bf 1}^n)\nonumber\\
	&=&f({\bf Q}_d{\bf Y}_d + {\bf U}_d{\bf Z}_d+{\bf b}_{\widetilde{X}_d} {\bf 1}^n)
\end{eqnarray}

Now we can formulate the desired component factorization into an optimization problem with the cost function:
\begin{eqnarray}
	\label{eqn:cost}\Omega^*=\mathop{argmin}_{\Omega}
	~\Delta({\bf Y}_r,{\bf Y}_d)
	&+&\lambda~\norm{\Omega}_2\nonumber\\
	+~\zeta_r~\Delta({\bf X}_r,\widetilde{\bf X}_r)
	&+&\zeta_d~\Delta({\bf X}_d,\widetilde{\bf X}_d)\nonumber\\
	+~\alpha_r~\Psi({\bf Y}_r;\rho_Y)
	&+&\alpha_d~\Psi({\bf Y}_d;\rho_Y)\nonumber\\
	+~\beta_r~\Psi({\bf Z}_r;\rho_Z)
	&+&\beta_d~\Psi({\bf Z}_d;\rho_Z)
\end{eqnarray}

where $\Omega=\{{\bf W}_., {\bf V}_., {\bf Q}_., {\bf b}_.\}$ is the set of all parameters, and $\left[\lambda,\zeta_.,\alpha_.,\beta_.\right]$ are hyper-parameters of trade-off between terms. 

The first term in (\ref{eqn:cost}) forces the shared components of the two modalities (${\bf Y}_r$ and ${\bf Y}_d$) to be as close as possible. 
We formulate this term as the Frobenius norm of the difference between two matrices:
\begin{equation}
	\label{eqn:corrterm}
	\Delta({\bf Y}_r,{\bf Y}_d)=\norm{{\bf Y}_r-{\bf Y}_d}_F
\end{equation}

The second term is the general weight regularization term, applied on the projection weights to prevent networks from overfitting training data. 

The reconstruction costs are represented as $\Delta({\bf X}_r,\widetilde{\bf X}_r)$ and $\Delta({\bf X}_d,\widetilde{\bf X}_d)$ to prevent the model from degeneration.
Here, we use Frobenius norm (the same as (\ref{eqn:corrterm})) of the reconstruction error for the reconstruction cost term.

Last four terms of (\ref{eqn:cost}) are sparsity penalty terms over ${\bf Y}$ and ${\bf Z}$ outputs. 
It has been shown in \cite{sparseAutoEncoder,marc2007efficient} that applying sparsity on the features of ${\bf Y}$ and $\bf Z$ will help to improve the learning capability, especially when components are overcomplete. 
As our sparsity penalty, we use KL-divergence term, applied between $\bf Y$ components and the sparsity parameters $\rho_Y$, as well as $\bf Z$ and $\rho_Z$ .

It is worth pointing out, since the proposed framework is built on a sparse autoencoder-like scheme and has sigmoid scaling nonlinearity, it is necessary to apply PCA whitening on the input matrices ${\bf X}_r$ and ${\bf X}_d$ and scale their elements into the range of $\left[0,1\right]$.

In this formulation, the disparity between $\bf Z_d$ and $\bf Z_r$ components is applied implicitly.
The similarity inducing norm pushes the common components of the two modalities to move inside $\bf Y$ components. 
Therefore, we expect the remaining features in each of $\bf Z$ components to be highly different across the modalities.
\begin{figure}[t]
	\centering
	\def\svgwidth{\columnwidth}
	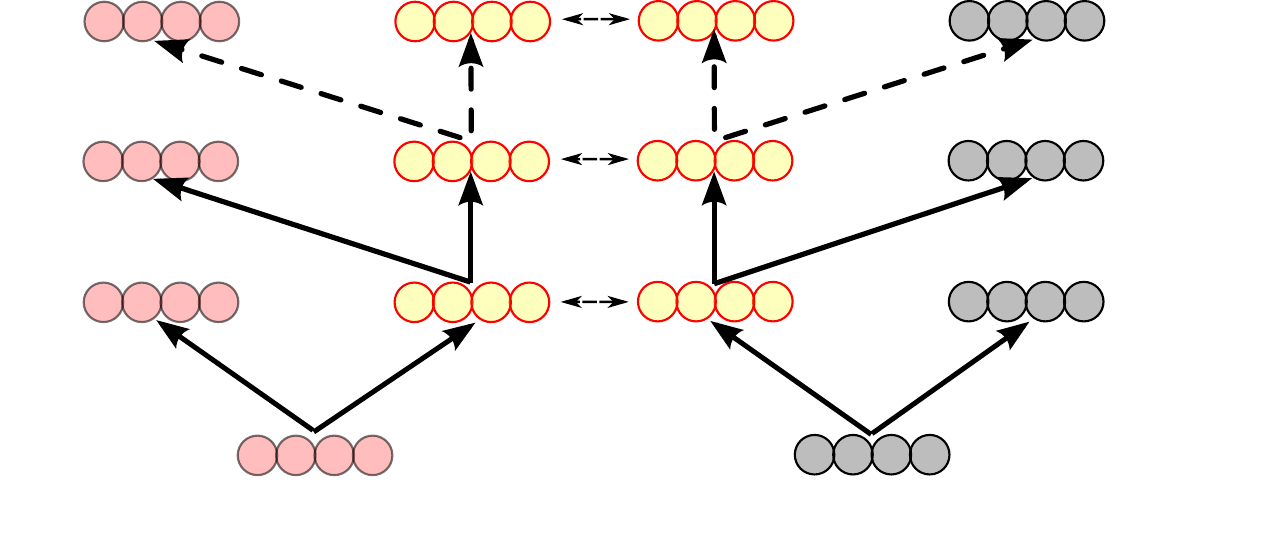
	\caption{Cascading factorization layers to a deep shared-specific network.
		To disentangle the highly nonlinear combination of shared-specific components, factorization layers are stacked by feeding the ${\bf Y}$ components of each layer as inputs of the next layer.}
	\label{fig:deep}
\end{figure}

\subsection{Deep shared-specific component analysis}

State-of-the-art RGB and depth based features for action recognition, are extracted by multiple linear and nonlinear layers of projection, embedding, spatial and temporal pooling, or statistical distribution encodings, \eg BOvW \cite{BOVW} and FV ~\cite{FV} or Fourier temporal pyramids in \cite{actionletPAMI}.
Hence the common components between modalities can lie on highly complex and nonlinear subspaces of input data, and one layer of the proposed shared-specific analysis cannot decode these complexities between the components.

By cascading multiple shared-specific analysis layers, we build a deep network to further factorize input features based on their higher orders of common and private information between modalities. 
To do so, we feed $\bf Y$ components of the previous layer as multimodal inputs of the current layer and apply the same method with new learning parameters in order to further factorize the features. 
As illustrated in \figurename{~\ref{fig:deep}}, each layer extracts modality-specific components of the modalities and passes the shared ones for further factorization in the next layer:
\begin{equation}
	\begin{bmatrix}
		{\bf Y}_r^{(i)}\\{\bf Y}_d^{(i)}\\{\bf Z}_r^{(i)}\\{\bf Z}_d^{(i)}
	\end{bmatrix}
	=
	\begin{cases}
		g({\bf X}_r,{\bf X}_d;\Omega^{(i)}) & \mbox{if } i=1 \\
		g({\bf Y}_r^{(i-1)},{\bf Y}_d^{(i-1)};\Omega^{(i)}) & \mbox{if } i>1
	\end{cases}
\end{equation}

By applying this hierarchy on nonlinear layers, we expect the network to factorize more complex and higher order components of the inputs as it moves forward through the layers.
Our deep network is trained greedily and layer-wise \cite{bengio2007greedy,hinton2006fast,Lee:2009:CDB:1553374.1553453}. 
In other words, the optimization of each layer is started upon the convergence of the previous layer's training.

Upon training of the deep network, each input sample will be factorized into a pair of specific components $({\bf Z}_r^{(i)},{\bf Z}_d^{(i)})$ for each layer $i\in\left[1,..,l\right]$, plus the concatenation of last layer's shared components $({\bf Y}_r^{(l)},{\bf Y}_d^{(l)})$.

\subsection{Convolutional shared-specific component analysis}
\label{sec:conv}

The input features (${\bf X}_r, {\bf X}_d$) of the proposed deep shared-specific component analysis network are assumed to describe different representations of a multimodal entity.
In our application, this entity is a RGB+D human action video and the inputs are RGB-based and depth-based features of it.

\begin{figure}[t]
	\centering
	\def\svgwidth{\columnwidth}
	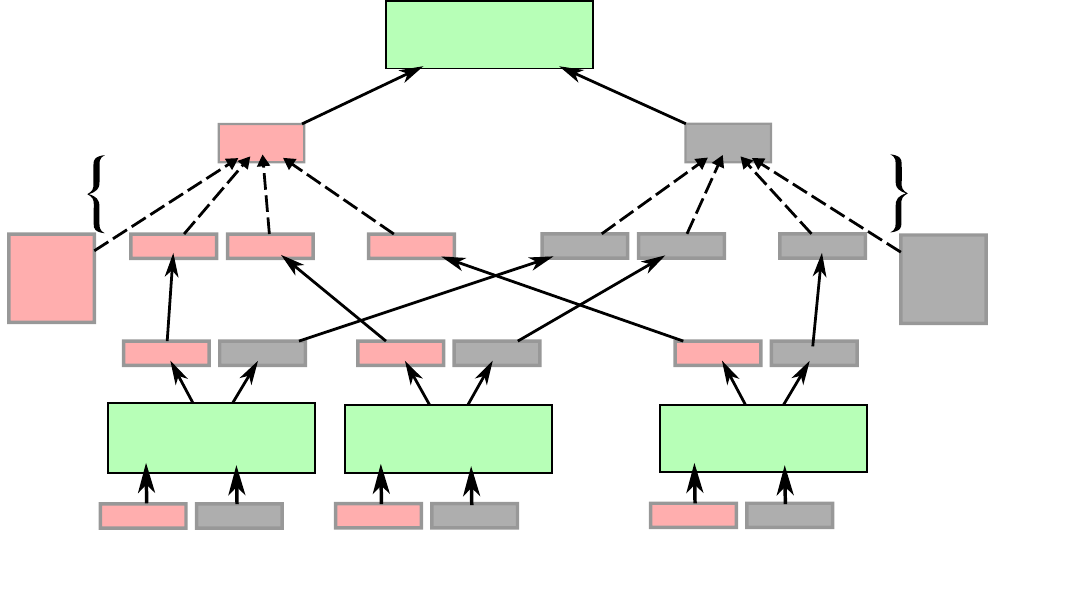
	\caption{Schema of our convolutional and holistic networks of deep shared-specific component analysis ($DSSCA$).
		We divide each video into $n$ local cubes.
		Local features ${\bf X}_r^i$ and ${\bf X}_d^i$ are extracted from the $i^{th}$ cube.
		Convolutional network (denoted as $DSSCA^L$) is trained and then applied to decompose local features.
		The factorized components are then combined with holistic features ${\bf X}_r^H$ and ${\bf X}_d^H$.
		This combination undergoes PCA and is fed into the holistic network (denoted as $DSSCA^H$) as its multimodal input.
	}
	\label{fig:conv}
\end{figure}

Since every input video can be regarded as a three dimensional cube (in $x, y, t$), it can be split to sub-cubes along all of its three dimensions, and the proposed multimodal analysis can be done on each of these sub-cubes separately.
By limiting our analysis into holistic RGB+D features, we may lose discriminative local information in both modalities,
because local features also have dependencies across modalities and their deep shared-specific component analysis (DSSCA) is beneficial.
Therefore, as depicted in \figurename{~\ref{fig:conv}}, we first train the local DSSCA network (DSSCA$^L$) on RGB+D features of the sub-cubes of training video samples.

One can think of this stage as applying the same DSSCA$^L$ network on all the sub-cubes of every input RGB+D video.
At each step, we have a fixed-sized window over the current sub-cube in both RGB and depth channels of the input video and feed their corresponding sub-video representations to the DSSCA$^L$ network as a single training sample.
By convolving this window over all of the possible sub-cubes of every input video sample, we train the DSSCA$^L$ network.

The learned DSSCA$^L$ is then utilized to decompose the multimodal features of all the convolved sub-cubes.
For every input video sample, we concatenate all of the factorized components of its sub-cubes.
The resulting representation is then put together with the holistic multimodal features of video sample, to build the input for the holistic DSSCA network (DSSCA$^H$), similar to~\cite{deepISACVPR2011}.
The inputs of DSSCA$^H$ are PCA whitened and scaled into the range of $\left[0,1\right]$.

Overall, we have $L=l_1+l_2$ layers of factorization where $l_1$ and $l_2$ are the number of layers in DSSCA$^L$ and DSSCA$^H$ networks respectively.
By applying the trained local-holistic networks into the features of each video sample, we have a set of $2L+1$ independent components:
\begin{equation}
	\label{eqn:allcomponenets}
	{\bf A} = \{{({\bf Z}_r^1)}^T,{({\bf Z}_d^1)}^T,...,{({\bf Z}_r^L)}^T,{({\bf Z}_d^L)}^T,{({\bf Y}^L})^T\}^T
\end{equation} 
where ${\bf Y}^L=\begin{bmatrix}{\bf Y}_r^L\\{\bf Y}_d^L\end{bmatrix}$ is the concatenation of last layer's common components.

\subsection{Optimization algorithm}
\label{sec:optimization}
The proposed formulation of cost function (\ref{eqn:cost}) is not a convex function of training parameters. 
Therefore, optimization of the learning parameters is not feasible in a single step. 
We iteratively optimize subsets of the parameters while keeping others fixed to achieve a suboptimal solution which is already shown effective in different applications \cite{5206757}. 

Specifically, the learning parameters of each layer can be divided into two subsets. 
First are the ones defined for projection and reconstruction of the shared components ${\bf Y}_.$, 
and second consists of similar parameters for individual component ${\bf Z}_.$. 
These two sets are:
\begin{equation}
	\Omega_{Y}=\{{\bf W}_r,{\bf W}_d,{\bf Q}_r,{\bf Q}_d,{\bf b}_{Y_r},{\bf b}_{Y_d},{\bf b}_{\widetilde{X}_r},{\bf b}_{\widetilde{X}_d}\}
\end{equation}
\begin{equation}
	\Omega_{Z}=\{{\bf V}_r,{\bf V}_d,{\bf U}_r,{\bf U}_d,{\bf b}_{Z_r},{\bf b}_{Z_d},{\bf b}_{\widetilde{X}_r},{\bf b}_{\widetilde{X}_d}\}
\end{equation}
Now, to optimize the overall cost, we first fix $\Omega_Z$ (except ${\bf b}_{\widetilde{X}_.}$) and minimize the cost function (\ref{eqn:cost}) regarding $\Omega_Y$. 
Then fix parameters of $\Omega_Y$ (except ${\bf b}_{\widetilde{X}_.}$) and optimize regarding $\Omega_Z$ and repeat this iteratively to converge into a suboptimal point.

In our implementation, all the optimization steps are done by ``L-BFGS'' algorithm using off-the-shelf ``minFunc'' software \cite{schmidt2005minfunc}.

\section{Structured sparsity learning machine}
\label{sec:classifier}

Previous shared-specific analysis steps were all unsupervised and applied just based on the mutual characteristics of the two modalities.
As a result, the factorized features of each component are not guaranteed to be equally discriminative for the following classification step.
Hence we adopt the structured sparsity regularization method of \cite{icml2013_wang13c,6619242} aiming to select a number of components/layers sparsely to achieve more robust classification.
Since the features of each component are highly correlated, our structured sparsity regularizer bounds the weights of the features inside each component to become activated or deactivated together.

Mathematically, we want to learn a linear projection matrix $\bf B$ to project our hierarchically factorized features ${\bf A}$ (see equation \ref{eqn:allcomponenets}), to a class assignment matrix $\bf F$ defined as:
\begin{equation}
	f_i^j=
	\begin{cases}
		1 & \mbox{if $j^{th}$ sample belongs to the $i^{th}$ class} \\
		0 & \mbox{otherwise}
	\end{cases}
\end{equation}
so that ${\bf A}^T{\bf B}$ would be as close as possible to $\bf F$. 

Each column of $\bf A$ consists of $2L+1$ components of features for each training sample. 
We use the notation ${\bf A}^G$ to denote the rows of $\bf A$ which include the features of component $G$. 
Variable $G$ can take values between 1 and $2L+1$ or their corresponding component labels. 
Correspondingly, columns of $\bf B$ have the same structure, and we denote the $G^{th}$ component's parameters as ${\bf B}^G$. 
We refer to the $i^{th}$ column of $\bf B$ as ${\bf b}_i$ which is the projection to our binary classifier for the $i^{th}$ action. 
Finally ${\bf b}_i^G$ refers to the $i^{th}$ column of ${\bf B}^G$.

Our classifier is formulated as another optimization problem with the cost function below. 
\begin{eqnarray}
	\label{eqn:cost2}
	{\bf B}^*&=&\mathop{argmin}_{{\bf B}}\norm{{\bf A}^T{\bf B}-{\bf F}}_F^2
	~+~\gamma_E~\norm{{\bf B}}_{G_E}\nonumber\\
	&+&\gamma_L~\norm{{\bf B}}_{G_L}
	~+~\gamma_W~\norm{{\bf B}}_F
\end{eqnarray}

Component-wise regularizer norm, $\|{\bf B}\|_{G_E}$, groups the weights of each component by applying a $\ell_2$ norm. 
Then applies the component selection by a $\ell_1$ norm over the $\ell_2$ values of all components. 
Mathematically:
\begin{eqnarray}
	\label{eqn:GEnorm}
	\norm{{\bf B}}_{G_E}&=&\sum\limits_{i=1}^c\sum\limits_{G=1}^{2L+1}\norm{{\bf b}_i^G}_2\nonumber\\
	&=&\sum\limits_{i=1}^c\sum\limits_{j=1}^L\left(\norm{{\bf 
			b}_i^{Z_r^j}}_2+\norm{{\bf b}_i^{Z_d^j}}_2\right)\nonumber\\
	&+&\sum\limits_{i=1}^c\norm{{\bf b}_i^{Y^L}}_2
\end{eqnarray}
where $c$ is the number of class labels. 

This mixed norm dictates the component-wise weight learning regarding their discriminative strength for each action class. 
Since it applies $\ell_2$ norm inside the components and $\ell_1$ norm between them, 
it regularizes the weights within each component, 
while sparsely selects discriminative components for different classes.

On the other hand, a layer-wise group selection can also be beneficial, because discriminative features may become factorized in some layers of our hierarchical deep network. 
Based on this intuition, we apply another group sparsity mixed norm to enforce layer selection. 
Similar to $G_E$ norm, our layer selection norm ($G_L$) groups the learning parameters corresponding to the components of each layer of the network, and applies $\ell_1$ sparsity between them:
\begin{eqnarray}
	\label{eqn:GLnorm}
	\norm{{\bf B}}_{G_L}=\sum\limits_{i=1}^c\sum\limits_{j=1}^L\begin{Vmatrix}
		\begin{bmatrix}{\bf b}_i^{Z_r^j}\\{\bf b}_i^{Z_d^j}\end{bmatrix}
	\end{Vmatrix}_2+\sum\limits_{i=1}^c\norm{{\bf b}_i^{Y^L}}_2
\end{eqnarray}

The last norm in (\ref{eqn:cost2}) is a general weight decay regularizer to prevent the entire classifier from overfitting.

Similar to previous section, this optimization is also done using ``L-BFGS'' algorithm. 
Upon training the classifier and finding the optimal ${\bf B}^*$, 
we classify each testing sample with exemplar features $\bf a_q$ as:
\begin{equation}
	h({\bf a}_q) = \mathop{argmax}_c~\langle~{\bf a}_q,{\bf b}_c^*~\rangle
\end{equation}

\section{CCA-RICA factorization as a baseline method}
\label{sec:ccarica}

As a baseline to the proposed method to perform the shared-specific analysis of the RGB+D inputs, we combined canonical correlation analysis (CCA) \cite{hotelling1936relations,CCASurvey} and reconstruction independent component analysis (RICA) \cite{RICA_NIPS11}, to extract correlated and independent components of input features. 
In this section we describe this baseline method.

We use the notation ${\bf X}_r$ to represent input local RGB features, and ${\bf X}_d$ for corresponding local depth features.
We define the linear projections of the two input features as: 
\begin{align}
	\label{eqn:CIA:YrYd}{\bf Y}_r={\bf W}_{r,c} {\bf X}_r \hspace{20pt},\hspace{20pt} {\bf Y}_d={\bf W}_{d,c} {\bf X}_d
\end{align}
and to make them maximally correlated we maximize:
\begin{align}
	\label{eqn:CIA:cca}\mathop{maximize}_{{\bf w}_{r,c}^j,{\bf w}_{d,c}^j} \hspace*{5pt} &Corr({\bf Y}_r^j,{\bf Y}_d^j)\nonumber\\
	= \hspace*{5pt} &Corr({\bf w}_{r,c}^j {\bf X}_r,{\bf w}_{d,c}^j {\bf X}_d)
\end{align}
in which superscript $j$ refers to the $j^{th}$ row of the corresponding matrices.

Canonical correlation analysis \cite{hotelling1936relations,CCASurvey} solves this analytically as an eigenproblem, 
in which each eigenvector gives one row of the projection and altogether provides the full projection matrices which lead to the maximum correlation between them.

Based on our intuition about insufficiency of shared components for recognition tasks, in the second step, we fix correlation projections (${\bf W}_{r,c} , {\bf W}_{d,c}$) and apply a reconstruction independent component analysis formulation \cite{RICA_NIPS11}, to extract modality-specific components for RGB and depth separately.
\begin{equation}
	\label{eqn:CIA:ZrZd}{\bf Z}_r={\bf W}_{r,i} {\bf X}_r \hspace{20pt},\hspace{20pt} {\bf Z}_d={\bf W}_{d,i} {\bf X}_d
\end{equation}

For RGB features we optimize:
\begin{align}
	\label{eqn:CIA:ricar}\mathop{mininize}_{{\bf w}_{r,i}} &~ \frac{\lambda}{m}
	\norm{
		\widetilde{\bf X}_r - {\bf X}_r
	}_F^2~+~
	\sum_{j}{\norm{{\bf W}_{r,i}^j {\bf X}_r}_1}\nonumber\\
	where &~ \widetilde{\bf X}_r = \left[{\bf W}_{r,c}^T , {\bf W}_{r,i}^T\right]
	\begin{bmatrix}
		{\bf W}_{r,c}\\
		{\bf W}_{r,i}
	\end{bmatrix}
	{\bf X}_r
\end{align}

Similarly for depth features we optimize:
\begin{align}
	\label{eqn:CIA:ricad}\mathop{mininize}_{{\bf w}_{d,i}} &~ \frac{\lambda}{m}
	\norm{
		\widetilde{\bf X}_d - {\bf X}_d
	}_F^2~+~
	\sum_{j}{\norm{{\bf W}_{d,i}^j {\bf X}_d}_1} \nonumber\\
	where &~ \widetilde{\bf X}_d = \left[{\bf W}_{d,c}^T , {\bf W}_{d,i}^T\right]
	\begin{bmatrix}
		{\bf W}_{d,c}\\
		{\bf W}_{d,i}
	\end{bmatrix}
	{\bf X}_d
\end{align}

Upon convergence of (\ref{eqn:CIA:ricar}) and (\ref{eqn:CIA:ricad}), the RGB+D features of each trajectory ($k$) can be represented as a quadruple: $\{{\bf Z}_r(k),{\bf Y}_r(k),{\bf Y}_d(k),{\bf Z}_d(k)\}$.

\section{Experiments}
\label{sec:exp}

This section presents our experimental setup and the results of the proposed methods on three RGB+D action recognition datasets.

\subsection{Experimental setup}
\label{sec:expsetup}
The proposed methods are evaluated on five RGB+D action recognition datasets.
All these datasets are collected using the Microsoft Kinect sensor in an indoor environment \cite{kinectSurvey2012}.
This sensor captures RGB videos and depth map sequences, and locates the 3D positions of 20 body joints of actors in the scene.

In our experiments, we try to use features which encode information regarding all the available modalities. 
From RGB videos, we extract dense trajectories \cite{idt_ICCV2013} and use HOG, HOF, MBHX, and MBHY features as trajectory descriptors. 
To encode the global representation of samples based on their trajectories, we use VQ with 2K codewords, for each descriptor.
The final representation of each sample video, is the concatenated max-pooled codes of the four descriptors, over 3 levels of the temporal pyramid. 
For depth sequences, we use the histograms of oriented 4D normals (HON4D) features \cite{HON4D}. 
To explore different setups on each dataset, we extract this feature in different settings. 
We describe the details in following subsections.

Since the RGB and depth sequences are not fully aligned and not synced in most of the datasets (all the evaluated ones in this paper, except RGBD-HuDaAct), convolutional cubes have to be large enough so that they mostly cover the same parts of the video between the two modalities. 
To apply the convolutional network, we consider four temporal quarters of the videos. 
In this way, each input sample has four temporal segments in our convolutional network and the factorized components of all these segments, together with holistic features of the entire sample are considered as the inputs of the stacked network. 

To cover various aspects of RGB+D motion and appearances of input samples, we used a combination of different features.
For depth channel, we extract Fourier coefficients of the joint locations and local occupancy pattern (LOP) features \cite{actionletCVPR}, 
histogram of oriented 4D normals (HON4D) \cite{HON4D}, dynamic skeletons (DS), and dynamic depth patterns (DDP) \cite{jianfang_CVPR15}.
From RGB videos, we extract dynamic color patterns (DCP) \cite{jianfang_CVPR15} and dense trajectory features \cite{idt_ICCV2013}.

For depth-based input, we use Fourier coefficients of the joint locations and local occupancy pattern (LOP) features \cite{actionletCVPR}.
The size of ${\bf X}_.$, ${\bf Y}_.$, and ${\bf Z}_.$ vectors is fixed as 100 for local features and 200 for holistic and stacked networks in our experiments.
On each of the experiments, the optimal values of gammas in SSLM are found via leave-one-sample-out cross-validation over training samples.

To show the effectiveness of our method, we compare it with two baseline methods below:

{\bf Baseline method 1}: descriptor level fusion.
In this method, we concatenate all the input RGB and depth-based features and train a linear SVM for classification.

{\bf Baseline method 2}: kernel level combination.
For this baseline method, we calculate the RBF kernel matrices based on all the input RGB and depth-based features and combine them linearly to classify in the form of multi-kernel SVM.
We find the weights of kernels via a brute force search in a cross validation setting using training samples \cite{Bosch:2007:RSS:1282280.1282340}.

In the following tables, we report the results of our method in two settings:

{\bf DSSCA Kernel} is the kernel combination of the hierarchically factorized components of our shared-specific analysis network.

{\bf DSSCA SSLM}: refers to the proposed structured sparsity learning machine based on the hierarchically factorized components described in section \ref{sec:classifier}.

It is worth mentioning, there are more than 40 datasets specifically for 3D human action recognition.
The survey of Zhang \etal \cite{RGB-D_Survey_40_Pichao} provided a great coverage over the current datasets and discussed their characteristics in different aspects, as well as the best performing methods for each dataset.

\subsection{Online RGBD action dataset}
\label{sec:online}
\begin{table}
	\setlength{\tabcolsep}{3pt} 
	\begin{center}
		\begin{tabular}{|c||c|c||c|c|}
			\hline
			Eval.  & Baseline 	& Baseline	& DSSCA  & DSSCA	\\
			Dataset	& Method 1	& Method 2 	& Kernel	& SSLM	\\
			\hline\hline
			Online S1	& 86.6\% 	& 91.1\% 	& 92.9\% 	& \bf 95.5\% \\
			\hline
			Online S2	& 85.6\%	& 91.0\% 	& 91.9\% 	& \bf 93.7\% \\
			\hline
			Online S3	& 73.0\% 	& 80.2\% 	& 82.0\% 	& \bf 83.8\% \\
			\hline		
		\end{tabular}
	\end{center}
	\caption{Comparison of the results of our methods with the baselines in Online RGBD Action dataset.
		S1, S2, and S3 refers to the three different scenarios of the Online RGBD Action dataset.
		First column shows the performance of descriptor concatenation on all RGB+D input features.
		Second column reports the accuracy of the kernel combination on the same set of features.
		Third column shows the result of our correlation-independence analysis. 
		It employs a kernel combination for classification.
		Last column reports the accuracy of proposed structured sparsity learning machine.}
	\label{tab:onlinevsbl}
\end{table}

\begin{table}
	\setlength{\tabcolsep}{3pt} 
	\begin{center}
		\begin{tabular}{|c|c||c|c|}
			\hline
			Evaluation &	Network 	& DSSCA	& DSSCA	\\
			Dataset & Structure	& Kernel	& SSLM		\\
			\hline\hline
			Online S1 & Holistic & 90.2\% 	& 92.0\% \\
			\hline
			Online S1 & Local  	& 92.9\% 	& 93.8\% \\
			\hline
			Online S1 & Stacked Local+Holistic	& 92.9\% 	& \bf 95.5\% \\
			\hline\hline
			Online S2 & Holistic	& 87.4\% 	& 91.0\%\\
			\hline
			Online S2 & Local 	& 88.3\% 	& 89.2\% \\
			\hline
			Online S2 & Stacked Local+Holistic	& 91.9\% 	& \bf 93.7\% \\
			\hline\hline
			Online S3 & Holistic	& 79.3\% 	& 82.0\% \\
			\hline
			Online S3 & Local 	& 75.7\% 	& 77.5\% \\
			\hline
			Online S3 & Stacked Local+Holistic	& 82.0\% 	& \bf 83.8\% \\
			\hline
		\end{tabular}
	\end{center}
	\vspace*{10pt}
	\caption{Performance comparison for holistic network, local network, and stacked local+holistic (\figurename{~\ref{fig:conv}}) networks on Online RGBD action datasets.
		Reported are the results of our method using kernel combination and SSLM.}
	\label{tab:onlineoverall}
\end{table}

Online RGBD action dataset \cite{Orderlet} is a RGB+D benchmark for action recognition.
Unlike most of the other RGB+D benchmarks, this dataset is collected in different locations and provides a cross-environment evaluation setting.
It includes samples of 7 daily action classes: \emph{drinking, eating, using laptop, reading cellphone, making phone call, reading book}, and \emph{using remote}.
For the recognition task, it provides videos of 24 actors.
Each actor performs each of the actions twice. 
Overall, this dataset include 336 RGB+D video samples.
Three different recognition scenarios are defined on this dataset.
The first and second scenarios are cross-subject tests. 
In the first scenario, the first 8 actors are assigned for training and the second 8 actors are for testing. 
The samples of the second scenario are the same as the first one but training and testing samples are swapped. 
The third scenario is a cross-environment setting. 
The videos of the third 8 actors are collected in another location and are considered as test data. 
The other 16 actors' videos are used for training.
The first and second scenarios are cross-subject and the third is a cross-environment evaluation.

\tablename{~\ref{tab:onlinevsbl}} compares the results of the deep shared-specific component analysis (DSSCA) and structure sparsity learning machine (SSLM), with baseline methods on this dataset.
The results of this experiment show our DSSCA network successfully decompose input features into a more powerful representation which leads into a clear improvement on the classification performance. 
They also show our SSLM can select the discriminative components and layers and learns a better classifier.

We also compare different structures of our DSSCA network.
For each scenario, we report the performance of three structures.
``Holistic'' refers to the 3-layer deep network applied on holistic features.
``Local'' is the 2-layer convolutional network applied on local features.
``Stacked local+holistic'' is the stacked local and holistic networks, as illustrated in \figurename{~\ref{fig:conv}}.
The results are reported in \tablename{~\ref{tab:onlineoverall}}.
We conclude that the local and holistic features are complementary and applying stacked local+holistic network can improve the final classification accuracy.

In our third experiment on this dataset, performance of the proposed networks is compared with a similar network without modality-specific components. 
The reference network acts similarly to traditional CCA methods.
We compare these two networks on the ``local'' network of third scenario.
The result is shown in \tablename{~\ref{tab:onlineZcompare}}.
We can see including independent components is beneficial and improves the accuracy.
Performance of the network with these components is clearly higher.
The second observation is our method improves the performance more significantly by having multiple layers.
Without having the modality-specific components, the values of common components can not change much, on higher layers.
This shows our proposed structure is suitable for cascading more layers and decomposing the features layer by layer.

\tablename{~\ref{tab:onlinecomp}} compares our results with the state-of-the-art method on this dataset.
Due to the recency of this dataset, only two other works reported results on this dataset.
As shown, our method outperforms their results with a large margin, which demonstrates the importance of RGB+D fusion for action recognition as well as the effectiveness of our proposed method for this task.

\begin{table}
	\begin{center}
		\begin{tabular}{|c||c|c|}
			\hline
			Network			& Layer 1	& 2 Layers\\
			Description 	& SSLM		& SSLM	\\
			\hline\hline
			Local Without $\bf Z$ & 73.0\% 	& 73.9\% \\
			\hline
			Local With $\bf Z$ & 76.6\% 	& \bf 77.5\% \\
			\hline
		\end{tabular}
	\end{center}
	\caption{Comparison with a correlation network (without modality-specific components) on the Online RGBD Action dataset, local network, scenario~3.
		Without $\bf Z$ components, the network is limited to the shared ones and acts similar to CCA.}
	\label{tab:onlineZcompare}
\end{table}

\begin{table}
	\begin{center}
		\begin{tabular}{|c|c|c|}
			\hline
			Methods &	Setup 	& Accuracy	\\
			\hline\hline
			HOSM \cite{HOSM} & Same environment & 49.5\% \\
			\hline
			Orderlet \cite{Orderlet} & Same env. & 71.4\% \\
			\hline
			Meng \etal \cite{7284883} & Same env. & 75.8\% \\
			\hline
			Proposed DSSCA-SSLM & Same env.	& \bf 94.6\%\\
			\hline\hline
			HOSM \cite{HOSM} & Cross env. & 50.9\% \\
			\hline
			Orderlet \cite{Orderlet} & Cross env.	& 66.1\%\\
			\hline
			Proposed DSSCA-SSLM & Cross env.	& \bf 83.8\%\\
			\hline
		\end{tabular}
	\end{center}
	\caption{Performance comparison of proposed DSSCA with the state-of-the-art results on Online RGBD Action dataset.
		Same environment setup is the average of S1 and S2 scenarios, and cross environment setup is the same as S3 scenario.}
	\label{tab:onlinecomp}
\end{table}

\subsection{MSR-DailyActivity3D dataset}
\label{sec:daily}
\begin{table}
	\setlength{\tabcolsep}{3pt} 
	\begin{center}
		\begin{tabular}{|c||c|c||c|c|}
			\hline
			Eval.  & Baseline 	& Baseline	& DSSCA  & DSSCA	\\
			Dataset	& Method 1	& Method 2 	& Kernel	& SSLM	\\
			\hline\hline
			Daily & 91.9\% 	& 94.4\% & 96.3\% 	& \bf 97.5\% \\
			\hline		
		\end{tabular}
	\end{center}
	\caption{Comparison of the results of our methods with the baselines in MSR-DailyActivity3D dataset.}
	\label{tab:dailyvsbl}
\end{table}

\begin{table}
	\setlength{\tabcolsep}{3pt} 
	\begin{center}
		\begin{tabular}{|c|c||c|c|}
			\hline
			Evaluation &	Network 	& DSSCA	& DSSCA	\\
			Dataset & Structure	& Kernel	& SSLM		\\
			\hline\hline
			Daily & Holistic & 95.0\%	& 96.3\% \\
			\hline
			Daily & Local 	& 95.0\% 	& 96.9\% \\
			\hline
			Daily & Stacked Local+Holistic	& 96.3\% 	& \bf 97.5\% \\
			\hline
		\end{tabular}
	\end{center}
	\vspace*{10pt}
	\caption{Performance comparison for holistic network, local network, and stacked local+holistic (\figurename{~\ref{fig:conv}}) networks on MSR-DailyActivity3D dataset.
		Reported are the results of our method using kernel combination and SSLM.}
	\label{tab:dailyoverall}
\end{table}

\begin{table}
	\setlength{\tabcolsep}{20pt} 
	\begin{center}
		\begin{tabular}{|c|c|}
			\hline
			Method & Accuracy\\
			\hline\hline
			HoDG-RDF \cite{6836044} & 74.5\%\\\hline
			Bag-of-FLPs \cite{7008115} & 78.8\%\\\hline
			HON4D \cite{HON4D} & 80.0\%\\\hline
			SSFF \cite{AmirAthens} & 81.9\%\\\hline
			ToSP \cite{6918467} & 84.4\%\\\hline
			RGGP \cite{RGGP} & 85.6\%\\\hline
			Actionlet \cite{actionletPAMI} & 85.8\%\\\hline
			SVN \cite{Yang_2014_CVPR} & 86.3\%\\\hline
			BHIM \cite{Kong_2015_CVPR} & 86.9\%\\\hline
			DCSF+Joint \cite{xiaCVPR13spatio} & 88.2\%\\\hline
			MMTW \cite{MMTW} & 88.8\%\\\hline
			HOPC \cite{HOPC_PAMI} & 88.8\%\\\hline
			Depth Fusion \cite{Yu_DepthFusion_ACM} & 88.8\%\\\hline
			MMMP \cite{MMMP_PAMI} & 91.3\%\\\hline
			DL-GSGC \cite{Luo_2013_ICCV} & 95.0\%\\\hline
			JOULE-SVM \cite{jianfang_CVPR15} & 95.0\%\\\hline
			Range-Sample \cite{RangeSample} & 95.6\%\\\hline\hline
			Proposed DSSCA-SSLM & \bf 97.5\%\\\hline
		\end{tabular}
	\end{center}
	\caption{Performance comparison of the proposed multimodal DSSCA with the state-of-the-art methods on MSR-DailyActivity dataset.}
	\label{tab:dailycomp}
\end{table}

MSR-DailyActivity dataset \cite{actionletCVPR} is among the most challenging RGB+D benchmarks for action recognition, which has a high level of intra-class variation and a large number of action classes. 
It provides 320 RGB+D samples, from 16 classes of daily activities: \emph{drink, eat, read book, call cellphone, write on a paper, use laptop, use vacuum cleaner, cheer up, sit still, toss paper, play game, lie down on sofa, walk, play guitar, stand up,} and \emph{sit down}. 
Each action is done by 10 actors, twice by each actor. 
The standard evaluation on this dataset is defined on a cross-subject setting: first five subjects are used for training and others for testing. 
Results of the experiments on this benchmark are reported in \tablename{s~\ref{tab:dailyvsbl} and \ref{tab:dailyoverall}. 
	
\tablename{~\ref{tab:dailycomp}} also shows the accuracy comparison between the proposed method and the state-of-the-art results reported on this benchmark, in which we reduced the error rate by more than 40\% compared to the best reported results so far.
This shows our RGB+D analysis method can effectively improve the performance of the action recognition system.

\subsection{3D action pairs dataset}
\label{sec:pairs}
\begin{table}
\setlength{\tabcolsep}{3pt} 
\begin{center}
	\begin{tabular}{|c||c|c||c|c|}
		\hline
		Eval.  & Baseline 	& Baseline	& DSSCA  & DSSCA	\\
		Dataset	& Method 1	& Method 2 	& Kernel	& SSLM	\\
		\hline\hline
		Pairs & 97.7\% 	& 98.3\% 	& \bf 100.0\% 	& \bf 100.0\% \\
		\hline		
	\end{tabular}
\end{center}
\caption{Comparison of the results of our methods with the baselines in 3D Action Pairs dataset.}
\label{tab:pairsvsbl}
\end{table}

\begin{table}
\setlength{\tabcolsep}{3pt} 
\begin{center}
	\begin{tabular}{|c|c||c|c|}
		\hline
		Evaluation &	Network 	& DSSCA	& DSSCA	\\
		Dataset & Structure	& Kernel	& SSLM		\\
		\hline\hline
		Pairs & Holistic & 98.9\% 	& 99.4\% \\
		\hline
		Pairs & Local 	& 99.4\% 	& 98.9\% \\
		\hline
		Pairs & Stacked Local+Holistic & \bf 100.0\% 	& \bf 100.0\% \\
		\hline
	\end{tabular}
\end{center}
\vspace*{10pt}
\caption{Performance comparison for holistic network, local network, and stacked local+holistic (\figurename{~\ref{fig:conv}}) networks on 3D Action Pairs dataset.
	Reported are the results of our method using kernel combination and SSLM.}
\label{tab:pairsoverall}
\end{table}

\begin{table}
\setlength{\tabcolsep}{20pt} 
\begin{center}
	\begin{tabular}{|c|c|}
		\hline
		Method & Accuracy\\
		\hline\hline
		DHOG \cite{DHOG} & 66.11\%\\\hline
		Bag-of-FLPs \cite{7008115} & 75.56\%\\\hline
		Actionlet \cite{actionletPAMI} & 82.22\%\\\hline
		HON4D \cite{HON4D} & 96.67\%\\\hline
		MMTW \cite{MMTW} & 97.22\%\\\hline
		HOG3D-LLC \cite{HOG3DLLC} & 98.33\%\\\hline
		HOPC \cite{HOPC_PAMI} & 98.33\%\\\hline
		SVN \cite{Yang_2014_CVPR} & 98.89\%\\\hline
		MMMP \cite{MMMP_PAMI} & 100.0\%\\\hline
		BHIM \cite{Kong_2015_CVPR} & 100.0\%\\\hline\hline
		Proposed DSSCA-SSLM & \bf 100.0\%\\
		\hline
	\end{tabular}
\end{center}
\caption{Performance comparison of proposed multimodal correlation-independence analysis with the state-of-the-art methods on 3D Action Pairs dataset.}
\label{tab:pairscomp}
\end{table}

3D Action Pairs dataset \cite{HON4D} is a less challenging RGB+D dataset for action recognition.
This dataset provides 6 pairs of action classes: \emph{pick up a box/put down a box, lift a box/place a box, push a chair/pull a chair, wear a hat/take off a hat, put on a backpack/take off a backpack,} and \emph{stick a poster/remove a poster}.
Each pair of the classes have almost the same set of body motions but in different temporal order.
Each action class is captured from 10 subjects, each one 3 times. 
Overall, this dataset includes 360 RGB+D video samples.
The first five subjects are kept for testing and others are for training. 

\tablename{~\ref{tab:pairscomp}} compares the accuracies between the proposed framework and the state-of-the-art methods reported on this benchmark.
Our method ties with two recent works (MMMP \cite{MMMP_PAMI}, and BHIM \cite{Kong_2015_CVPR}) in saturating the benchmark by achieving the flawless $100\%$ accuracy on this dataset.

\subsection{NTU RGB+D dataset}
\label{sec:nturgbd}

NTU RGB+D \cite{Amir-Dataset-CVPR} is one of the largest scale benchmark dataset for 3D action recognition.
It provided 56880 RGB+D video samples of 60 distinct actions.
The 60 action classes in NTU RGB+D dataset are: \emph{
drinking,
eating,
brushing teeth,
brushing hair,
dropping,
picking up,
throwing,
sitting down,
standing up (from sitting position),
clapping,
reading,
writing,
tearing up paper,
wearing jacket,
taking off jacket,
wearing a shoe,
taking off a shoe,
wearing on glasses,
taking off glasses,
puting on a hat/cap,
taking off a hat/cap,
cheering up,
hand waving,
kicking something,
reaching into self pocket,
hopping,
jumping up,
making/answering a phone call,
playing with phone,
typing,
pointing to something,
taking selfie,
checking time (on watch),
rubbing two hands together,
bowing,
shaking head,
wiping face,
saluting,
putting palms together,
crossing hands in front.
sneezing/coughing,
staggering,
falling down,
touching head (headache), 
touching chest (stomachache/heart pain),
touching back (back-pain),
touching neck (neck-ache), 
vomiting,
fanning self.
punching/slapping other person,
kicking other person,
pushing other person,
patting other's back,
pointing to the other person,
hugging,
giving something to other person,
touching other person's pocket,
handshaking,
walking towards each other, and
walking apart from each other.}

\begin{table}
\setlength{\tabcolsep}{3pt} 
\begin{center}
	\begin{tabular}{|c||c||c|}
		\hline
		Eval.  & Baseline	& DSSCA	\\
		Dataset	& Method 1 	& SSLM	\\
		\hline\hline
		NTU RGB+D & 59.7\% 	& \bf 74.9\% \\
		\hline		
	\end{tabular}
\end{center}
\caption{Comparison of the result of our method with the baseline for the cross-subject evaluation criteria of NTU RGB+D dataset.}
\label{tab:nturgbdvsbl}
\end{table}

\begin{table}
\setlength{\tabcolsep}{3pt} 
\begin{center}
	\begin{tabular}{|c|c||c|}
		\hline
		Evaluation &	Network 	& DSSCA	\\
		Dataset & Structure	& SSLM		\\
		\hline\hline
		NTU RGB+D & Holistic & 70.4\% \\
		\hline
		NTU RGB+D & Local 	& 66.4\% \\
		\hline
		NTU RGB+D & Stacked Local+Holistic & \bf 74.9\% \\
		\hline
	\end{tabular}
\end{center}
\vspace*{10pt}
\caption{Performance comparison for holistic network, local network, and stacked local+holistic (\figurename{~\ref{fig:conv}}) networks on the cross-subject evaluation criteria of NTU RGB+D dataset.
	Reported are the results of our method using SSLM.}
\label{tab:nturgbdoverall}
\end{table}

\begin{table}
\begin{center}
	\begin{tabular}{|l|c|}
		\hline
		Method  & Cross-Subject Accuracy\\
		\hline\hline
		HOG$^2$ \cite{hog2-ohnbar}					& 32.24\% \\\hline
		Super Normal Vector \cite{Yang_2014_CVPR}	& 31.82\% \\\hline
		HON4D \cite{HON4D} 							& 30.56\% \\\hline
		
		Lie Group \cite{VemulapalliCVPR14}			& 50.08\% \\\hline
		Skeletal Quads \cite{skeletalQuads}			& 38.62\% \\\hline
		FTP Dynamic Skeletons \cite{jianfang_CVPR15}& 60.23\% \\\hline
		HBRNN-L \cite{rnnskeleton_cvpr15}			& 59.07\% \\\hline
		P-LSTM \cite{Amir-Dataset-CVPR}				& 62.93\% \\\hline
		ST-LSTM \cite{Liu_2016_ECCV}				& 69.20\% \\\hline\hline
		Proposed DSSCA - SSLM 			& \bf 74.86\%\\\hline
	\end{tabular}
\end{center}
\caption{Performance comparison of proposed multimodal correlation-independence analysis with the state-of-the-art methods on the cross-subject evaluation criteria of NTU RGB+D dataset.}
\label{tab:nturgbdcomp}
\end{table}

Unlike other evaluated datasets, NTU RGB+D is collected by Microsoft Kinect v.2.
Therefore, its skeletal data includes more body joints and is more accurate.
For our experiments in this section, we limit the depth-based features to Fourier temporal pyramids over skeletons, HON4D and LOP. For RGB-based inputs we use the same set of features used for the other datasets.

This dataset suggested two evaluation criteria, cross-subject and cross-view.
For the cross-view evaluation, our set of RGB based features perform very poorly and could not contribute powerful enough in the proposed multimodal analysis.
Therefore, we evaluate the proposed DSSCA-SSLM framework only on the cross-subject criterion of this dataset.

Due to the large size of training video samples in this dataset, evaluation of the kernel-based methods (both baseline method 2 and DSSCA-kernel) were not tractable and we only reported the results for baseline method 1 and DSSCA-SSLM frameworks, as provided in \tablename{s \ref{tab:nturgbdvsbl}} and \ref{tab:nturgbdoverall}.
\tablename{~\ref{tab:nturgbdcomp}} compares the performance of the proposed framework in comparison with other state-of-the-art on this benchmark.

\subsection{RGBD-HuDaAct Dataset}

RGBD-HuDaAct \cite{rgbdhudaact} is a large size benchmarks for human daily action recognition in RGB+D.
This dataset includes 1189 RGB+D video sequences from 13 action classes: \emph{exit the room, make a phone call, get up from bed, go to bed, sit down, mop floor, stand up, eat meal, put on jacket, drink water, enter room, take off jacket,} and \emph{background activity}.
The standard evaluation on this dataset is defined on a leave-one-subject-out cross-validation setting.
In our experiments we follow the evaluation setup described in \cite{rgbdhudaact}.
\begin{table}
\setlength{\tabcolsep}{3pt} 
\begin{center}
	\begin{tabular}{|c||c|c||c|c|}
		\hline
		Eval.  & Baseline 	& Baseline	& DSSCA  & DSSCA	\\
		Dataset	& Method 1	& Method 2 	& Kernel	& SSLM	\\
		\hline\hline
		HuDaAct & 95.1\% & 97.6\% 	& 98.3\% 	& \bf 99.0\% \\
		\hline		
	\end{tabular}
\end{center}
\caption{Comparison of the results of our methods with the baselines on RGBD-HuDaAct dataset.
	First column shows the performance of descriptor concatenation on all RGB+D input features.
	Second column reports the accuracy of the kernel combination on the same set of features.
	Third column shows the result of our correlation-independence analysis. 
	It employs a kernel combination for classification.
	Last column reports the accuracy of proposed structured sparsity learning machine.}
\label{tab:hudavsbl}
\end{table}

\begin{table}
\setlength{\tabcolsep}{3pt} 
\begin{center}
	\begin{tabular}{|c|c||c|c|}
		\hline
		Scenario &	Network 	& DSSCA	& DSSCA	\\
		Number & Structure	& Kernel	& SSLM		\\
		\hline\hline
		HuDaAct & Holistic & 98.3\% 	& \bf 99.0\% \\
		\hline
		HuDaAct & Local & 98.7\% & 98.7\% \\
		\hline
		HuDaAct & Stacked Local+Holistic & 98.3\% & \bf 99.0\% \\
		\hline
	\end{tabular}
\end{center}
\caption{Performance comparison for holistic network, local network, and stacked local+holistic networks on RGBD-HuDaAct dataset.
	Reported are the results of our method using kernel combination and SSLM.}
\label{tab:hudaoverall}
\end{table}

\begin{table}[!t]
\renewcommand{\arraystretch}{1.3}
\begin{center}
	\begin{tabular}{|c|c|}
		\hline
		\bfseries Method & \bfseries Accuracy\\\hline
		3D-MHIs \cite{rgbdhudaact} & 70.5\%\\\hline
		iM$^2$EDM \cite{chatzis2013infinite} & 76.8\%\\\hline
		MF-HMM \cite{Kosmopoulos2013} & 78.6\%\\\hline
		DLMC-STIPs \cite{rgbdhudaact} & 81.5\%\\\hline
		DIMC-STIPs \cite{depthinduced} & 87.7\%\\\hline
		STIP HOGHOF+LDP \cite{6411953} & 89.1\%\\\hline
		Part-based BOW-Pyramid \cite{6696669} & 91.7\%\\\hline
		RGB+D Linear Coding \cite{Liu201579} & 92.0\%\\\hline
		CCA (Atomic Level) & 93.9\%\\\hline
		CCA-RICA (Atomic Level) & 96.4\%\\
		\hline\hline
		{\bf Single Unit SSCA (Atomic Level)} & {\bf 97.9\%}\\\hline
		{\bf DSSCA-SSLM (Global Level)} & {\bf 99.0\%}\\\hline
	\end{tabular}
\end{center}
\caption{Performance Comparison on RGBD-HuDaAct Dataset}
\label{tab:hudacomp}
\end{table}

\subsubsection{Atomic Local Level Feature Analysis}
Unlike most of the other datasets, this benchmark provides fully synchronized and aligned set of RGB and depth videos.
This important characteristic enables us to apply the atomic level of analysis on local RGB and depth features within the video samples.

As our atomic local level features, we extract the tracked dense trajectories \cite{idt_ICCV2013} in RGB sequences and their HOG, HOF, MBHX, and MBHY descriptors from both modalities.

To evaluate the effectiveness of the proposed RGB+D analysis, we apply a single layer SSCA to decompose RGB and depth descriptors of the trajectories to their correlated and independent components.
For training stage, we sample a set of 40K trajectories from training set.
The output of the analysis, which are four factorized components for each trajectory are clustered separately by K-Means with codebook size 1K.
LLC coding \cite{LLC} and BOF framework are applied on the codes of all the trajectories from each RGB+D video sample to extract their global representations.

In the final step, a linear SVM is used as the action classifier trained on the extracted global representations of the action video samples.

We evaluated the performance of canonical correlation analysis (CCA) method also.
As a better baseline, we also evaluated added independent components.
In our implementation of the CCA-RICA method (section \ref{sec:ccarica}) we used the provided codes by the authors of \cite{borga2001canonical} for CCA and \cite{RICA_NIPS11} for RICA.

All the optimizations in our experiments, are done using ``L-BFGS'' algorithm.
We use the off-the-shelf ``minFunc'' software released by \cite{schmidt2005minfunc}.

\tablename{~\ref{tab:hudacomp}} shows the results of all the experiments described in this section and compares them with other state-of-the-art methods.

At first, we evaluated the performance of correlated components of CCA without any modality specific features, which achieves 93.9\% outperforming all the reported results on this benchmark.
Compared to the accuracy of RGB+D linear coding \cite{Liu201579}, which has the most similar pipeline of action recognition to ours, CCA components shows about two percents improvement.
This approves the robustness of shared components and their advantage over using a simple combination of features from the two modalities.

In the next step, we apply RICA to extract modality-specific components for RGB and depth local features.
Adding specific components improves the accuracy of the classification by 2.5 more percents.
This supports our argument about the importance of modality-specific components and their discriminative strengths for action classification.
The confusion matrix for this method is illustrated in \figurename{~\ref{fig:confusion_ccarica}}.
The majority of the misclassification are caused by the background activity class.
This class contains samples of random motion and other simple activities which are not covered by other 12 classes, like walking around or stay seated without much of motion.
Therefore it is inevitable to have some confusion between this class with classes which contain very small amount of clear motion \eg making a phone call.
Similar action classes with reverse temporal order are also mixed up, \eg sit down and stand up, or put on jacket and take off jacket classes have the same appearance within individual frames, and their only differences are the arrangement of frames over time.

Next, we evaluate the proposed SSCA method on this atomic local level.
SSCA outperforms all other techniques by performing 97.9\% of correct classification and achieves the state-of-the-art accuracy on this dataset.
Compared to CCA-RICA method, SSCA improves the error rate by more than 40\% which is a notable improvement.
The confusion matrix of this experiment is also reported in \figurename{~\ref{fig:confusion_simssca}}.
Compared to the mixed-up cases of the CCA-RICA method (\figurename{~\ref{fig:confusion_ccarica}}), the confusion patterns are similar but furthered improved. 

\begin{figure}
\centering
\includegraphics[trim={1.7cm 18.8cm 3.0cm 1.8cm},clip,width=240pt]{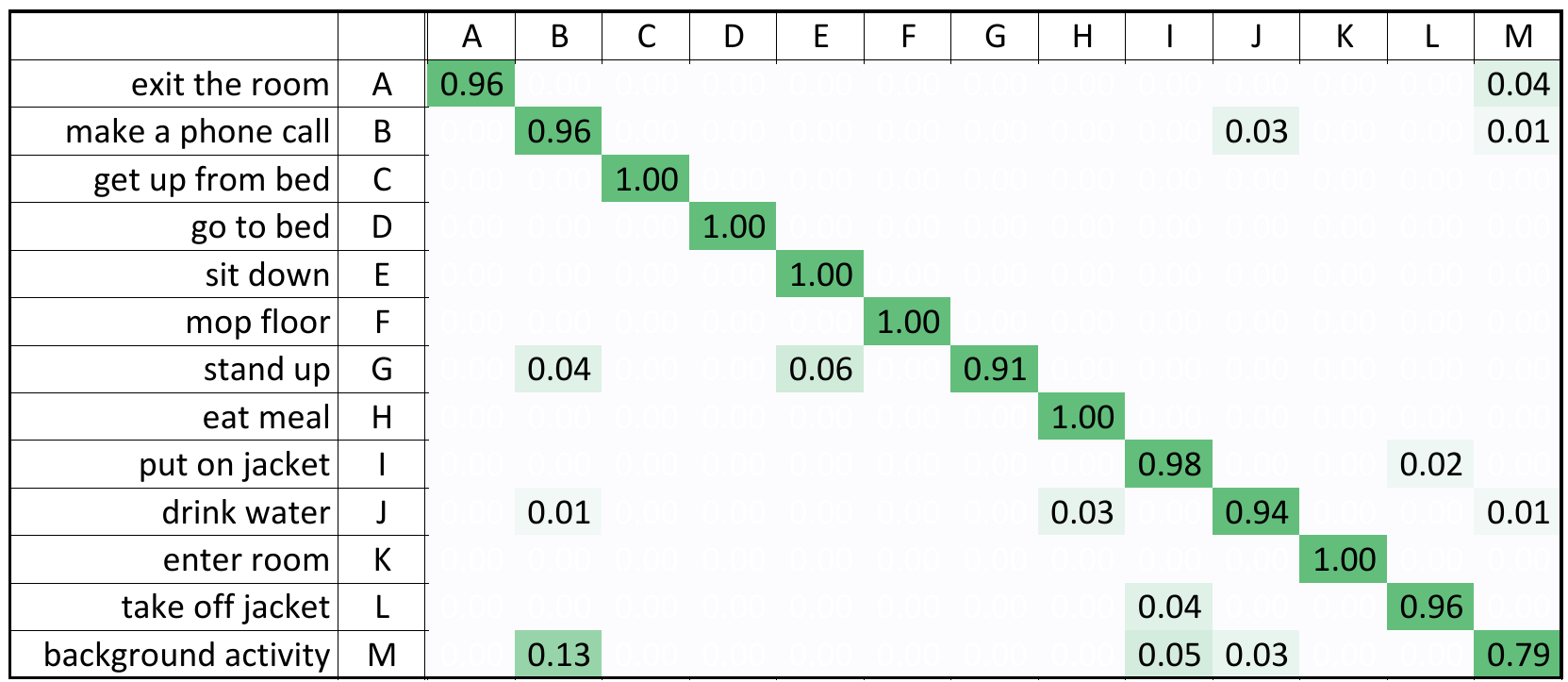} \\
\caption{Confusion matrix for CCA-RICA method on atomic local level features RGBD-HuDaAct dataset.
	Ground truth action labels are on rows and detections are on columns of the grid.}
\label{fig:confusion_ccarica}
\end{figure}

\begin{figure}
\centering
\includegraphics[trim={1.7cm 18.8cm 3.0cm 1.8cm},clip,width=240pt]{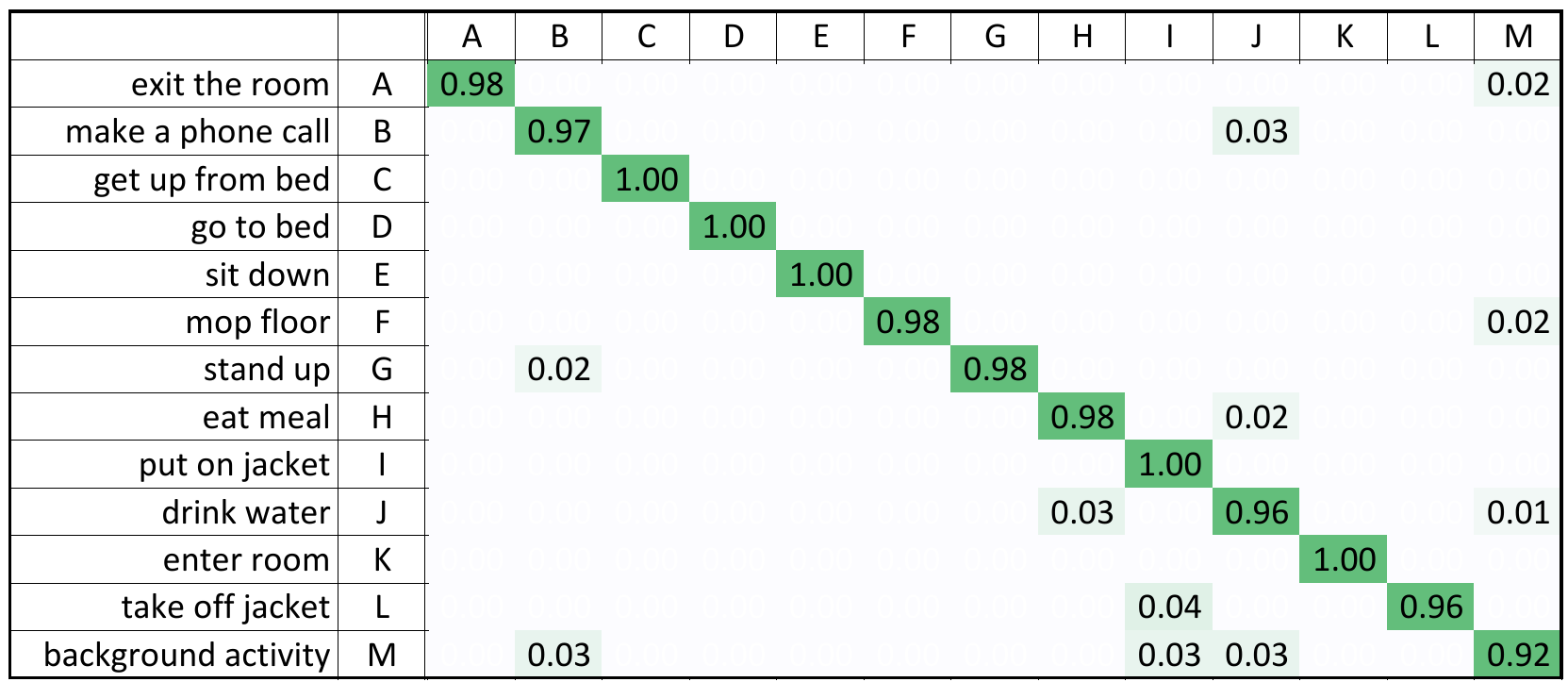} \\
\caption{Confusion matrix for SSCA method on atomic local level features of RGBD-HuDaAct dataset.
	Ground truth action labels are on rows and detections are on columns of the grid.}
\label{fig:confusion_simssca}
\end{figure}

\subsubsection{Global Level Feature Analysis}
Similar to other datasets reported in the paper, we perform the proposed RGB+D analysis on the global representations extracted from input samples.
For RGB signals, the features are HOG, HOF, MBHX, and MBHY descriptors of dense trajectories \cite{idt_ICCV2013}, followed by a K-means clustering and locality-constrained linear coding (LLC) \cite{LLC} to calculate their global representations as bags-of-features.
For depth, we extract HON4D features \cite{HON4D} for holistic and local depth based features.
The results of this experiment are reported in \tablename{s \ref{tab:hudavsbl} and \ref{tab:hudaoverall} in a similar evaluation setup to other datasets.

As can be seen in \tablename{~\ref{tab:hudacomp}}, applying DSSCA analysis in a deep and stacked framework outperforms all the current methods as well as the atomic local level analysis, and achieved the outstanding performance of 99.0\% on this benchmark, which shows more than 50\% improvement on the error rate compared to the atomic local level SSCA analysis.

Other reported results are also in accord with our results on other datasets and approve our arguments about the effectiveness of the the proposed framework.

\subsection{Comparison with single modality}
In \tablename{~\ref{tab:singlemodality}}, we compare our method with baseline method 2, based on single modality features.
Since each modality also has holistic and multiple local features, we perform baseline kernel combination to produce the results.
For a fair comparison, we use kernel combination for classification based on our factorized components.
It is not surprising to observe our method outperforms the baseline, since ours integrates RGB and depth information effectively.

\begin{table*}
	\setlength{\tabcolsep}{2.5pt} 
	\begin{center}
		\begin{tabular}{|c|c|c|c|c|c|c|}
			\hline
			Method 	& MSR Daily & 3D Action & RGBD & Online & Online & Online \\
			Method 	& Activity 3D	& Pairs& HuDaAct	& RGBD S1& RGBD S2& RGBD S3\\
			\hline\hline
			Baseline 2 on RGB-based &89.4\%&97.7\%&95.2\%&81.3\%&85.6\%&75.7\%\\
			Local+Holistic &&&&&&\\
			\hline
			Baseline 2 on depth-based &92.5\%&97.7\%&79.1\%&85.7\%&84.7\%&66.7\%\\
			Local+Holistic&&&&&&\\
			\hline\hline
			Ours Kernel& \bf 96.3\%&\bf 100.0\%&\bf 98.3\%&\bf 92.9\%&\bf 91.9\%&\bf 82.0\%\\
			\hline
		\end{tabular}
	\end{center}
	\caption{Comparison between our method and baseline method 2 on single modality RGB and depth based input features, on all the datasets.}
	\label{tab:singlemodality}
\end{table*}

\subsection{Analysis of component contributions in the classifier}
\tablename{ \ref{tab:componentanalysis}} shows the proportion of the weights assigned by SSLM to the factorized components of the stacked local+holistic networks.
The weights of ${\bf Y}^3$ are relatively high, which supports our initial argument about robustness and discriminative properties of the shared factorized components.
The Z components of the both modalities in all three layers also gain weights, which shows they also carry informative features and are complementary for the classification.
The reported values in this table shows how discriminative are the factorized features inside each of these components.
As can be seen, some components achieve very low (close to zero) values.
They hold important components in the distribution of the input multimodal data regardless of the action labels.
However, regarding the action classification task, they don't have considerable correlation with action labels and cannot contribute very much in classification, so they gain very low weights.

\begin{table}
	\setlength{\tabcolsep}{3.5pt} 
	\begin{center}
		\begin{tabular}{|c||c|c|c|c|c|c|c|}
			\hline
			Dataset  & ${\bf Z}^1_r$ & ${\bf Z}^2_r$ & ${\bf Z}^3_r$ & ${\bf Y}^3$ & ${\bf Z}^3_d$ & ${\bf Z}^2_d$ & ${\bf Z}^1_d$\\
			\hline
			Online S1	& 0.12 & 0.13 & 0.18 & 0.20 & 0.13 & 0.05 & 0.18 \\
			\hline
			Online S2	& 0.29 & 0.06 & 0.03 & 0.42 & 0.06 & 0.11 & 0.03 \\
			\hline
			Online S3	& 0.14 & 0.12 & 0.06 & 0.26 & 0.13 & 0.00 & 0.28 \\
			\hline
			Daily		& 0.22 & 0.05 & 0.07 & 0.23 & 0.08 & 0.08 & 0.28 \\
			\hline
			Pairs 		& 0.06 & 0.02 & 0.16 & 0.42 & 0.01 & 0.03 & 0.29 \\
			\hline			
		\end{tabular}
	\end{center}
	\caption{Proportion of the weights to factorized components in SSLM classifier for Online RGBD, MSR-DailyActivity3D, and 3D action pairs datasets.
		Reported values are the $\ell_2$ norms of all the corresponding weights to each of the components, learned by SSLM on the stacked local+holistic networks.}
	\label{tab:componentanalysis}
\end{table}

\section{Conclusion}
\label{sec:conclusion}
This paper presents a new deep learning framework for a hierarchical shared-specific component factorization (DSSCA), to analyze RGB+D features of human action videos.
Each layer of the proposed network is an autoencoder based component factorization unit, which decomposes its multimodal input features into common and modality-specific parts.
We further extended our deep factorization framework by applying it in a convolutional setting.

In addition, we proposed a structured sparsity based classifier (SSLM) which utilizes mixed norms to apply component and layer selection for a proper fusion of decomposed feature components.

Provided experimental results on five RGB+D action recognition datasets show the strength of our deep shared-specific component analysis and the proposed structured sparsity learning machine by achieving the state-of-the-art performances on all the reported benchmarks.


%

%


\section*{Acknowledgments}

This research was carried out at the Rapid-Rich Object Search (ROSE) Lab at the Nanyang Technological University, Singapore.  
The ROSE Lab is supported by the National Research Foundation, Singapore, under its Interactive Digital Media (IDM) Strategic Research Programme.

{\small
\bibliographystyle{ieee}
\bibliography{IEEEabrv,AmirPhdCon}
}

\end{document}